# A Survey on Lexical Ambiguity Detection and Word Sense Disambiguation

Miuru Abeysiriwardana
*School of Computing*
*Informatics Institute of Technology*
Colombo 06, Sri Lanka
miuru.2019450@iit.ac.lk

Deshan Sumanathilaka
*Department of Computer Science*
*Swansea University*
Swansea, Wales, United Kingdom
deshankoshala@gmail.com

*Abstract*— This paper explores techniques that focus on understanding and resolving ambiguity in language within the field of natural language processing (NLP), highlighting the complexity of linguistic phenomena such as polysemy and homonymy and their implications for computational models. Focusing extensively on Word Sense Disambiguation (WSD), it outlines diverse approaches ranging from deep learning techniques to leveraging lexical resources and knowledge graphs like WordNet. The paper introduces cutting-edge methodologies like word sense extension (WSE) and neuromyotonic approaches, enhancing disambiguation accuracy by predicting new word senses. It examines specific applications in biomedical disambiguation and language-specific optimisation and discusses the significance of cognitive metaphors in discourse analysis. The research identifies persistent challenges in the field, such as the scarcity of sense-annotated corpora and the complexity of informal clinical texts. It concludes by suggesting future directions, including using large language models, visual WSD, and multilingual WSD systems, emphasising the ongoing evolution in addressing lexical complexities in NLP. This thinking perspective highlights the advancement in this field to enable computers to understand language more accurately.

## I. Introduction

The field of Natural Language Processing (NLP) has seen groundbreaking improvements, particularly with the advent of deep learning and transformer-based models. These models, capable of learning complex patterns in language by training on large amounts of data, have greatly enhanced our ability to deal with the subtleties and complexities of language ambiguity [1]. Language ambiguity is a persistent challenge in NLP because human language can have multiple meanings depending on the context of a single word or phrase [1]. This is where the evolution of NLP comes into play, from rule-based systems, which relied on manually coded rules, to modern, sophisticated models. These modern models leverage machine learning algorithms to learn from large amounts of data, reflecting a variety of approaches. Each approach comes with its unique challenges and achievements. For instance, while transformer-based models have shown impressive performance on various tasks, and to train them effectively, a substantial amount of data and computational resources are required, which can be prohibitive for many applications. Despite these advancements, language ambiguity remains a complex problem, driving ongoing research and development in NLP technologies. This ongoing research and development in NLP are focusing more on the refinement of these models, targeting the creation of more intelligent, context-aware, and reliable systems, with promising results in enhancing the reasoning capabilities [2], marking a significant step in Artificial intelligence (AI) with the promise of more natural human-computer interactions in the years to come. As we make further progress in enhancing and perfecting these models, we can anticipate even more remarkable breakthroughs in NLP.

### A. Language Ambiguity

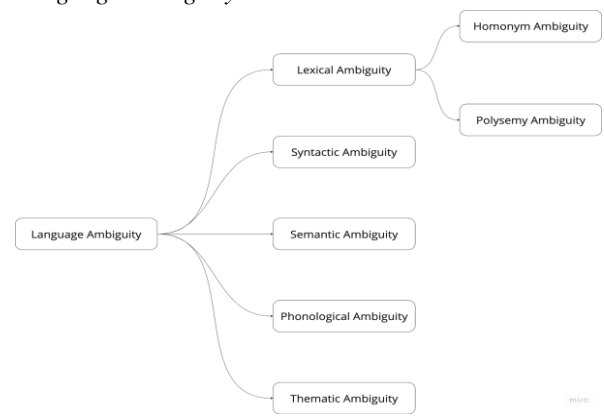

Fig. 1. Types of Ambiguity [1]

The language is a complex and constantly evolving system, frequently including words with multiple meanings. This poses significant challenges in the field of natural language processing (NLP), where the primary target is to facilitate computers in comprehending and interpreting human language with precision. Fig. 1 explores the broader topic of language ambiguity, which serves as a classification system for the complexities of language that can present challenges in communication and understanding. It is particularly relevant in the field of natural language processing, where disambiguation is a critical task. The major problem this paper intends to tackle is lexical ambiguity, where a given word can have various interpretations, making it especially difficult for computational models to process and understand the language effectively.

### B. Lexical Ambiguity

In Lexical ambiguity, homonymy, refers to a string of sounds or characters corresponding to more than one word and/or meaning[3]. Lexical ambiguity can be confusing in a sentence, as a word can have multiple meanings. For example, the word "green" in the sentence "That horse is green" can mean untrained or the color green [3]. Lexical ambiguity and syntactic ambiguity are two distinct concepts. Lexical ambiguity refers to the presence of multiple meanings for a single word or phrase. On the other hand, syntactic ambiguity occurs when a sentence or phrase can be interpreted in multiple ways due to its structure or grammar

[3]. For instance, take the more popular sentence, "I saw the man with a telescope." This sentence can be interpreted in two distinct ways: it might suggest that the narrator observed a man who was holding a telescope, or it could imply that the narrator used a telescope to see a man. This example clearly illustrates syntactic ambiguity, where the sentence structure leads to multiple interpretations. However, it's important to note that this differs from lexical ambiguity, which involves ambiguity in the meaning of individual words or phrases, not the overall sentence structure [1].

Polysemy refers to a single word with multiple meanings [4], [5]. The multiple meanings are listed under one entry in a dictionary [4]. An example of polysemy is the word 'dish' [6]. It can be seen in dictionaries that the word 'dish' has multiple definitions or polysemous meanings. The entry, "It's your turn to wash the dishes," refers to a kind of plate [4]. In terminological contexts, homonymy refers to the phenomenon where a single form represents different yet interconnected concepts [7]. This is distinct from homonymy in broader literary language, where it typically manifests as varied polysemantic formations lacking shared semantic elements [7].

The writer in [7] categorises terminological homonymy into three distinct forms: brunch, intersystem, and inter-scientific [7]. These categories highlight how a single term can possess multiple interpretations in different contexts or frameworks [7]. Take, for instance, the term "bank" in English, which is an example of a homonym. Depending on the situation, "bank" might refer to a financial establishment for depositing or borrowing money, or it could mean the area adjacent to a water body [6]. The same word, "bank", represents these two unrelated ideas, thereby qualifying as a homonym [7]. In NLP, detecting and resolving such lexical ambiguities are vital for enhancing the effectiveness of various technologies, including machine translation, information retrieval, sentiment analysis, and conversational AI [1]. The capacity of these systems to accurately interpret context and clarify word meanings is fundamental to their precision and dependability [8].

*C. Word Sense Disambiguation*

Determining the precise interpretation of words in context, Word Sense Disambiguation (WSD) stands as a central obstacle in the realm of Natural Language Processing. It delves beyond simply recognising ambiguous meanings, instead pinpointing and applying the appropriate sense of a word within its usage [8]. Consider the term "mouse" in a sentence like "A mouse involves an item operated by hand, equipped with one or more switches" – here, it would be inferred to mean the computer accessory [9]. While multiple methodologies have been examined within this domain, notably for English, the progress of WSD studies for South Asian tongues, such as Urdu, is still in a nascent phase [9].

Recently, various deep learning techniques have been applied with notable success to tasks in Natural Language Processing. A particular study scrutinised multiple deep learning strategies, such as basic recurrent neural networks, long short-term memory networks, gated recurrent units, bidirectional short-term memory, and ensemble learning approaches, for their effectiveness in Word Sense Disambiguation (WSD) for the Urdu language [8]. Using two standard corpora to gauge performance, this research revealed that these deep learning techniques surpassed previous achievements for the comprehensive WSD challenge in the Urdu language [8]. Additionally, a different method involved the integration of definitions from dictionaries to bolster the WSD function in neural language models [10]. This technique proved to be adaptable to various tasks with minimal need for extra parameters. Upon being assessed across 15 subsequent tasks, its performance was superior to other advanced methods on the SemEval and Senseval datasets and marked an improvement over the baseline results on the GLUE benchmark [10].

Neil Aspinall served as the Chief Executive of Apple Corps Ltd., a company distinct from Apple Inc., known for its technology products, and Apple Leisure Group, a travel and hospitality conglomerate and the fruit. Fig. 2. illustrates the process of assigning senses to the term 'Apple,' which is inherently ambiguous, with the most prominent interpretations including the companies and the common fruit. The primary objective of a Word Sense Disambiguation (WSD) system is to accurately determine the intended meaning of 'Apple' in each sentence by analysing surrounding contextual cues.

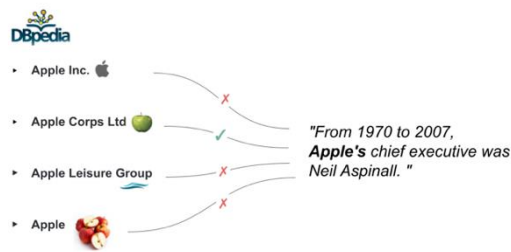

Fig. 2. Examples of WSD [11]

Fig 3 contains the functional WSD architecture proposed by [12].

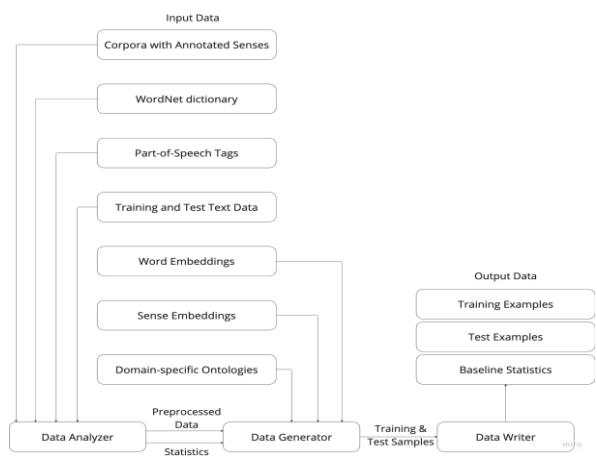

Fig. 3. WSD Functional Architecture [12]

The procedure begins with the system importing a dictionary derived from WordNet along with the training and testing textual materials into the Data Analyzer component. This component then transforms the imported data into a distinct intermediary structure and simultaneously computes elementary statistics pertinent to sentences and words [12]. These statistical analyses are instrumental in aiding users in determining the parameters needed to form training and test instances generated by the system. The Data Generator component, an essential part of the system, allows users to selectively extract diverse attributes from the training text data according to their set parameters. This component also produces training and test instances by amalgamating these attributes following the user's preferred sequence and structure. Following this, the created instances are transferred to the Data Writer component. This component is tasked with storing these instances as files for training and testing in a user-specified format and includes various baseline accuracy metrics to evaluate the WSD classifier [12]. These files are subsequently ready for use in external systems designed to learn and model classification patterns for the comprehensive all-words WSD task [12].

## II. RELATED WORK

### A. Word Sense Extention

A paradigm shift in NLP is observed in the concept of word sense extension (WSE). Instead of traditional disambiguation, this approach allows words to spawn new senses in novel contexts [13]. WSE creates a framework that blends cognitive models with learning schemes by dividing a word with multiple meanings into pseudo-tokens, each representing a different sense [13]. This method has demonstrated potential in forecasting likely new meanings for English words, particularly in transformer-based WSD models that handle infrequent word senses [13].

### B. Neurosymbolic Methodology for WSD Systems

Another breakthrough in WSD is achieved through a novel neurosymbolic methodology [14], surpassing the 80% F1 score ceiling and reaching 90%. This methodology integrates supervised learning with symbolic reasoning, utilizing hypernym relations and pre-trained word embeddings [15]. Such neurosymbolic approaches have opened new vistas for full-fledged multilingual WSD systems [15]. A recent study in Scientific Reports introduces a novel technique for creating word sense embeddings [16], especially for words that are polysemous. This method employs a bidirectional long short-term memory neural network and a self-attention mechanism. [16], this method enhances the accuracy of word sense induction and the quality of sense embeddings [16]. This advancement significantly contributes to the representation of word senses in computational linguistics. These developments reflect the dynamic nature of research in lexical ambiguity detection and WSD. Incorporating novel methodologies like WSE and neurosymbolic approaches, along with advancements in word sense embeddings, highlights the ongoing evolution and increasing sophistication in handling the complexities of language in NLP. These contributions enhance the understanding of lexical ambiguity and pave the way for more accurate and contextually relevant NLP applications.

### C. Tailoring Contextual Embeddings for WSD

In addition, there are multiple other exciting prospects in the field of WSD research. One notable example is the study by [17], which introduces a technique for WSD that involves tailoring contextual embeddings specifically for this task, relying exclusively on lexical information. The central concept of this method involves narrowing the semantic gap between related senses and contexts while distancing dissimilar or unrelated ones. This strategy has demonstrated superior results over past methods that modify contextual embeddings. Furthermore, it has reached a higher performance in knowledge-based WSD, particularly when combined with a reranking heuristic incorporating the sense inventory [17].

### D. Neurosymbolic Approaches for Reasoning within Graph Structures

Additionally, the study by [14] offers an extensive review of neurosymbolic approaches for reasoning within graph structures. [14] introduces an innovative categorisation system that allows for the classification and discussion of the primary applications of these methods. They also suggest various potential paths for the future development of this emerging research area [14].

### E. AMuSE-WSD: A Multilingual System for WSD

The "All-in-one Multilingual System for Easy Word Sense Disambiguation" (AMuSE-WSD) marks a notable breakthrough in the field of WSD [18]. Recently, WSD has garnered increased interest thanks to the impressive capabilities of deep learning methods in this area, particularly when enhanced by advanced pre-trained language models. Previously, there was a lack of comprehensive, user-friendly systems for leveraging these advancements, posing a challenge for researchers. AMuSE-WSD fills this void as the inaugural comprehensive system delivers top-tier sense information across 40 languages via a cutting-edge neural WSD model [18]. This innovation is pivotal for embedding semantic understanding into practical applications and spurring further research in lexical semantics. Before AMuSE-WSD, users looking to apply WSD in downstream tasks had to depend on existing systems mostly based on graph-centric heuristics or traditional machine-learning techniques. AMuSE-WSD revolutionises this by offering a leading-edge neural model for WSD [18]. By providing quality sense information in various languages, it is a significant resource for researchers and professionals in the field. This advancement underscores research's dynamic and evolving nature in detecting lexical ambiguity and advancing WSD, illustrating the growing complexity and refinement in processing language within Natural Language Processing (NLP). Such advancements are crucial in deepening the understanding of lexical ambiguity and set the stage for NLP applications that are more precise and context sensitive.

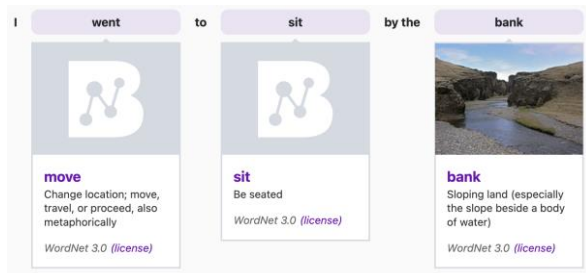

Fig. 4. Example 1 from AMuSE-WSD[18]

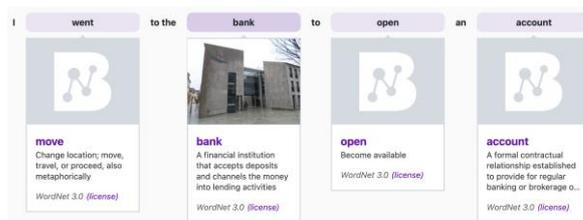

Fig. 5. Example 2 from AMuSE-WSD[18]

The sentences "I went to sit by the bank" and "I went to the bank to open an account" were used in the AMuSE-WSD system [18], as presented in Fig 4. And Fig 5. The system could accurately identify the intended reference of the word "bank" by considering the context. In the first sentence, "bank" refers to the edge of a river, while in the second, it refers to a financial institution. This demonstrates the system's ability to understand the nuances of language and interpret words based on their contextual meaning. It's a significant advancement in the field of NLP, highlighting the system's effectiveness in handling language ambiguity.

## III. FINDINGS AND ANALYSIS

Deep Learning for Urdu WSD [9]: The study explores a range of deep learning architectures to resolve the complexities of discerning word senses in Urdu. It examines a suite of techniques, including gated recurrent units, long-short-term memory networks, bidirectional long-short-term memory networks, simple recurrent neural networks, and ensemble learning strategies, as explained in Fig 6. where 4 deep learning methods are involved [9], for a selected ambiguous word, the 4 given models would assign the senses, and the final predicted sense is dependent upon a vote from derived outputs from these 4 models, and if they all predict different senses, then the final decision would be the sense assigned by the deep learning model with the highest accuracy. The study found that modern deep learning models are better at accurately determining word meanings in Urdu for the All-Words WSD task compared to earlier techniques [9].

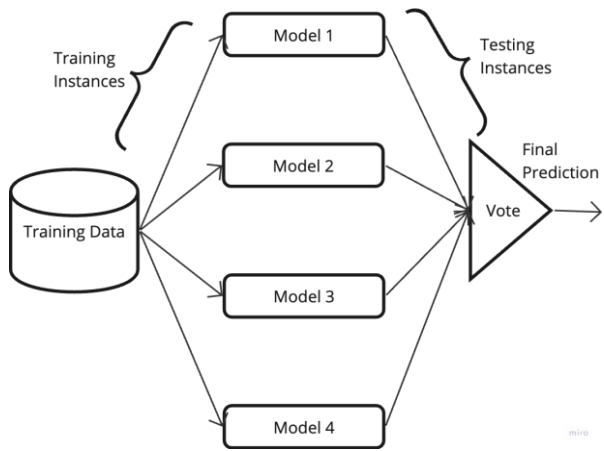

Fig. 6. URDU WSD Ensemble Learning Approach [9]

Biomedical Term Clarity Using Attention Neural Networks: This method introduces the use of attention neural networks to refine the precision in clarifying biomedical terminology [19]. It leverages the structure of words, grammatical roles, and contextual semantic cues from surrounding words to inform its features, achieving a notable average accuracy rate of 91.38% in distinguishing word senses [19]. Analysis of Metaphors in Cognition and Discourse: The study explores cognitive and discourse analysis of metaphors, examining the application of computer-aided tools such as Word Sense Disambiguation in detecting metaphorical language within discursive exchanges [20].

Employing the WordNet Knowledge Graph in WSD A novel approach employing the WordNet knowledge graph has been proposed to improve Word Sense Disambiguation (WSD) efforts. This method utilizes enhancements to the Lesk algorithm to better manage exact word matches and concise definitions, examines semantic similarity measures' relevance in WSD, and underscores three heuristic models: Most Frequent Sense, one sense per discourse, and one sense per collocation [21].

Optimizing Association Rules for WSD: The article presents a novel approach using an artificial immune system to refine association rules applicable to WSD [22]. It introduces a statistically normalized algorithm tailored for the Afaan Oromo language, designed to distinguish between meanings of ambiguous words independently of established rules or pre-labelled data. The algorithm exhibits encouraging outcomes, achieving an F-measure accuracy of 80.76% [22].

TABLE I. COMPARATIVE ASSESSMENT OF EXISTING WORKS

| Approach | Goals | Established Tools | Limitations |
|---|---|---|---|
| Investigation of deep learning methods. [9] | To outperform previous results for the Urdu All-Words WSD task. | Ensemble Learning, Gated Recurrent Units, Bidirectional Long-Short Term Memory, Recurrent Neural Networks, Long-Short Term Memory. | data requirements and computational complexity |
| Attention neural network-based method. [19] | Improve disambiguation accuracy of biomedical words. | Morphology, part of speech, and semantic information features. | challenges in integrating diverse feature sets |
| Computer-assisted analysis of metaphors in discourse. [12]. | Identify metaphorical features in discursive communication. | WSD algorithms. | detecting nuanced metaphorical usage. |
| Use of the WordNet knowledge graph. [13] | Novel approach for WSD | WordNet knowledge graph | limitations of WordNet's coverage and structure. |
| Artificial immune system for optimizing association rules.[22] | Perform WSD for the Afaan Oromo language without predefined rules or annotated datasets. | Normalized statistical algorithm. | Language-specific challenges and the generalizability of the method. |

Table I. presents a comprehensive overview of diverse approaches in computational linguistics and natural language processing, each characterized by unique goals, methodologies, and challenges. These vary from deep learning techniques targeted to enhance Urdu alphabet recognition to innovative uses of neural networks for biomedical text disambiguation, from metaphor analysis in discourse to WordNet for word sense disambiguation (WSD). A recurrent theme across the above approaches includes the challenges and the subtleties of natural language.

TABLE II. MODEL AND DATASET COMPARISON

| Approach | Model(s) Used | Dataset Used | Language | Accuracy | F1 Score |
|---|---|---|---|---|---|
| Investigation of deep learning methods. [9] | Ensemble of Simple RNN, LSTM, GRU, Bi-LSTM | ULS-WSD-18, UAW-WSD-18 | Urdu | 63.25% (UAW-WSD-18), 72.63% (ULS-WSD-18) | 0.66 |
| Resolving Amharic Lexical Ambiguity using Neural Word Embedding. [23] | joint supervised and unsupervised approach based on distributional semantic space for Amharic word sense disambiguation | Ambiguous Words, Amharic Annotated Corpora, Dictionary/Gloss | Amharic | 70% (supervised) 60% (unsupervised) 86% (joint) | 82.3% (supervised) 85.7% (unsupervised) 92.5% (joint) |
| Disambiguating spatial prepositions [24] | Bidirectional Encoder Representation from Transformer-based XLNet transformer | natural language expressions annotated for geo-spatial sense detection | English | 0.96 precision | 0.94 |
| Artificial immune system for optimizing association rules.[22] | Normalized Statistical Algorithm | Afaan Oromo ambiguous words | Afaan Oromo | - | 80.76% |

Table II. contains the dataset and model comparison of four different WSD approaches. Each method showcases the application of advanced NLP techniques to disambiguate word meanings in various languages, targeting different levels of resource availability and linguistic complexity.

## IV. CHALLENGES AND FUTURE DIRECTIONS

### A. Challenges in Lexical Ambiguity Detection and WSD

A semantic similarity-based method for WSD is developed using training sets from resources like SemCor, OMSTI, and the Adaptive-Lexical Resource [25]. The latter is a training compilation for supervised learning that focuses on words with multiple meanings [25]. A major challenge presents itself with polysemous words, where their interpretation is heavily context dependent. Training, therefore, necessitates focusing on the contextual nuances of these ambiguous terms [25]. Another problem that arises is related to the evaluation of uncertainty [26], [27]. Uncertainty evaluation is critical in various machine learning deployments, especially when safety is paramount. Current approaches to assess the calibration of uncertainty in regression are inadequate, as they fail to effectively separate useful uncertainty predictions from those that are not [27]. Additionally, there is a scarcity of large, manually annotated sense corpora for WSD [28], [29]. For instance, SemCor, the most extensive manually annotated sense corpus available, tags words with 33,760 different senses, representing only about 16% of WordNet's sense inventory [28]. There are efforts to mitigate this limitation by developing new sense-annotated corpora through automatic or semi-automatic means [29].

Significant challenges are associated with the informalities of clinical texts [30], such as typographical errors, inconsistent and incomplete texts, and non-standardized texts. The difficulty with the lack of clinical resources, such as clinical sense inventories, and privacy concerns for conducting research with clinical text are also notable [30].

### B. Future Directions

Utilisation of Large Language Models (LLMs) increases when enhancing phrases and clarifying ambiguities associated with specific words as knowledge bases [31]. An emerging area of interest lies in investigating cutting-edge transformer-based techniques for multimodal information retrieval [31]. The task of Visual Word Sense Disambiguation (VWSD) presents a unique challenge in selecting the most appropriate image from several options to accurately reflect the meaning of an ambiguous word in its context [32]. This task is currently being examined as a single-modal issue, transforming it into text-to-text, image-to-image retrieval, and question-answering formats to maximise the potential of the relevant models [32].

In the area of uncertainty estimation, future efforts are focused on developing new definitions that overcome existing shortcomings, alongside evaluating these using straightforward histogram-based methods [27]. Discussions are also underway regarding the practical constraints of existing methods in critical real-world scenarios, such as mission-critical and safety-sensitive applications [27]. Regarding Word Sense Disambiguation (WSD), there is a shift towards more refined results through adaptive word embeddings, particularly when these embeddings are crafted based on the context of the word [25].

The trend for Multilingual Word Sense Disambiguation (MWSD) is to create knowledge-driven and supervised MWSD systems [33]. Efforts are being made to develop unified sense representations that span multiple languages [33]. Additionally, the challenge of limited annotations in MWSD is being addressed by transferring knowledge from languages with abundant resources to those with fewer resources.

## V. CONCLUSION

The study on Word Sense Disambiguation (WSD) in natural language processing discusses several challenges, including the complexity of training models for context-dependent polysemous words and the limited availability of sense-annotated corpora. Addressing these, new methods, such as adaptive word embedding and the development of new automated and semi-automated corpora, are emerging. The field is advancing with Large Language Models (LLMs) for phrase enhancement and ambiguity resolution and exploring novel areas like Visual Word Sense Disambiguation (VWSD) for image-based context interpretation. Future directions also include refining uncertainty estimation in machine learning, which is crucial for safety-critical applications, and expanding into Multilingual Word Sense Disambiguation, focusing on unified sense representations and knowledge transfer between languages. These evolving strategies show the continuous progression in tackling lexical complexities in NLP, enhancing the capability for more accurate, context-aware language processing.